 \documentclass[accepted]{uai2022} 

\usepackage[american]{babel}

\usepackage{natbib} 
\usepackage{mathtools} 
\usepackage{booktabs} 
\usepackage{tikz} 
\usepackage[utf8]{inputenc} 
\usepackage[T1]{fontenc}    
\usepackage{hyperref}       
\usepackage{url}            
\usepackage{booktabs}       
\usepackage{amsfonts}       
\usepackage{nicefrac}       
\usepackage{microtype}      
\usepackage{xcolor}         
\usepackage{microtype}
\usepackage{booktabs} 
\usepackage[utf8]{inputenc} 
\usepackage[T1]{fontenc}    
\usepackage{hyperref}       
\usepackage{url}            
\usepackage{amsfonts}       
\usepackage{nicefrac}       
\usepackage{amsmath}
\usepackage{amsthm}
\usepackage{float}
\DeclareMathOperator*{\argmax}{argmax}
\usepackage{graphicx}

\title{Variational Interpretable Learning from Multi-view Data}

\author[1,3]{\href{mailto:<lq666988@gmail.com>?Subject=Your DICCA paper}{Lin Qiu\footnote{}}{}} 
\author[2]{Lynn Lin}
\author[1,3]{Vernon M. Chinchilli}
\affil[1]{%
    Department of Statistics\\
   The Pennsylvania State University
}
\affil[2]{%
    Department of Biostatistics and Bioinformatics\\
    Duke University
}
\affil[3]{%
    Department of Public Health Sciences\\
    The Pennsylvania State University
  }
  
\begin{document}
\maketitle

\begin{abstract} 
The main idea of canonical correlation analysis (CCA) is to map different views onto a common latent space with maximum correlation. We propose a deep interpretable variational canonical correlation analysis (DICCA) for multi-view learning.  The developed model extends the existing latent variable model for linear CCA to nonlinear models through the use of deep generative networks. 
DICCA is designed to disentangle both the shared and view-specific variations for multi-view data. To further make the model more interpretable, we place a sparsity-inducing prior on the latent weight with a structured variational autoencoder that is comprised of view-specific generators. Empirical results on real-world datasets show that our methods are competitive across domains.
\end{abstract}

\section{Introduction}
CCA \citep{Hotelling36} is a popular two-view data analysis technique which extracts the common information between two multivariate random variables by projecting them into a space in which they are maximally correlated. CCA has been used as a standard unsupervised two-view learning \citep{andrew13,wang-multi-view15}, a cross-view classification \citep{chang18,chandar16,kan16}, a representation learning on multiple views for prediction \citep{sargin07,dorfer16}, and for a classification from a single view when a second view is available \citep{arora12}.

When the data comes from several different views or modalities of the same underlying source of variation, the representation learning of CCA should be extended to the multi-view scenario. Since all views jointly characterize the same phenomenon, we can consider that there is shared information amongst them (or amongst subsets of them) by which they are all related. Further, we also expect that there might exist unique or private variations in each view, i.e., information which is specific to a single data stream. Variations that are not common to all dimensions are challenging to model. If included in the representation, a variation only present in a subset of dimensions will contaminate the representation of the dimensions that do not share this characteristic.

 A variety of extensions of CCA have been developed to learn a shared low-dimensional feature space of multi-view data, like kernel CCA \citep{lai00,akaho01}, shared kernel information embedding \citep{bach02}.  To capture the nonlinearity in complex data, deep neural network CCA (DCCA) was proposed \citep{andrew13}. DCCA is further extended to deep CCA autoencoder (DCCAE) \citep{wang-multi-view15} to improve the representation learning over DCCA by leveraging autoencoders to additionally reconstruct the inputs through reconstruction error terms for the objective function. While DCCA learns embeddings that capture shared variation, it does not explicitly model view-specific noise as in PCCA \citep{Bach05}. Recently, two deep variational CCAs (VCCA) were proposed \citep{tang17,wang16} which yield a generative two-view model with shared and view-specific factors.



Another important consideration for applying CCA for multi-view learning is the model interpretability, which is critical to understand systems of interaction among complex data. Lasso and group lasso are commonly used for simple interpretable models \citep{tibshirani94,yuan06}. They work by shrinking many model parameters toward zero and have seen great success in regression models, covariance selection \citep{danaher14}, linear factor analysis \citep{hirose12}, and group factor analysis \citep{Klami15}. Commonly, sparsity-inducing penalties are considered in the convex optimization literature due to their computational tractability using proximal gradient descent.

Latent variable modelling has been used widely for providing interpretable descriptions of data. A generative model is quite popular to achieve a compact representation through exploiting dependency structures in the observed data. The latent probabilistic version of CCA \citep{ghahramani15} is attractive in medical applications where the data are typically of small sample sizes but large feature spaces. However, the generative interpretation of CCA will ignore nonlinear structure in complex data such as images.

We present a more general deep interpretable CCA generative model (DICCA), in which the linear probabilistic layers are extended to deep generative multi-view networks. DICCA captures the variations of the views by a shared latent representation that describes most of the variability of multi-view data, and a set of view-specific factors. 
Our main contributions can be summarized as follows:

\begin{itemize}
    \item We propose a novel generative framework for CCA which can disentangle the shared and view-specific variations from multi-view data.
    \item We leverage the sparsity-inducing hierarchical Bayesian priors on the latent weight variables for an interpretable understanding on the feature level.
    \item We evaluate our approach on real datasets to demonstrate that our algorithm achieves better performance and yields an interpretable learning compared to state-of-the-art methods.
\end{itemize}

\section{Background}
CCA is a classical subspace learning method that aims to find the shared signal between a pair of data sources, by maximizing the sum correlation between corresponding projections. Let $\mathbf{X}_1 \in \mathbb{R}^{N\times d_1}$ and $\mathbf{X}_2 \in \mathbb{R}^{N\times d_2}$ be a pair of random vectors from two different views with $n$ independent samples and $d_1$, $d_2$ features, respectively. For ease of presentation, $\mathbf{X}_1$ and $\mathbf{X}_2$ are assumed to have null expectations. CCA maximizes the correlation between 
$\mathbf{a}_1 = \mathbf{u}_1^\top \mathbf{X}_1$ and $\mathbf{a}_2 = \mathbf{u}_2^\top \mathbf{X}_2$, where $\mathbf{u}_1$ and 
$\mathbf{u}_2$ are projection vectors. The first set of canonical directions is found via maximization of
\begin{equation*}
  \argmax_{\mathbf{u}_1,\mathbf{u}_2} \text{corr}\big(\mathbf{u}_1^\top \mathbf{X}_1, \mathbf{u}_2^\top \mathbf{X}_2\big)
  \label{eqn_cca}
\end{equation*}
and subsequent projections are found by maximization of the same correlation but in orthogonal directions. 
Combining the projection vectors into matrices $\mathbf{U}_1 = [\mathbf{u}_1^{(1)},\ldots,\mathbf{u}_1^{(k)}]$ and $\mathbf{U}_2 = [\mathbf{u}_2^{(1)},\ldots,\mathbf{u}_2^{(k)}]$ ($k \leq \text{min}(d_1,d_2)$), CCA can be reformulated as a trace maximization under orthonormality constraints on the projections, i.e.,
\begin{equation}
  \argmax_{\mathbf{U}_1,\mathbf{U}_2} \text{tr}( \mathbf{U}_1^\top \boldsymbol{\Sigma}_{12} \mathbf{U}_2 ) ~~~~ \text{s.t. } \mathbf{U}_1^\top \boldsymbol{\Sigma}_{1} \mathbf{U}_1 = \mathbf{U}_2^\top \boldsymbol{\Sigma}_{2} \mathbf{U}_2 = \mathbf{I},
  \label{eqn_cca_trace}
\end{equation}
for covariance matrices $\boldsymbol{\Sigma}_1=E(X_1 X_1^T$), $\boldsymbol{\Sigma}_2=E(X_2 X_2^T$), and cross-covariance matrix 
$\boldsymbol{\Sigma}_{12}=E(X_1 X_2^T$).
Let $\mathbf{T} = \boldsymbol{\Sigma}_{1}^{-1/2} \boldsymbol{\Sigma}_{12} \boldsymbol{\Sigma}_{2}^{-1/2}$ and its singular value decomposition (SVD) be $\mathbf{T} = \mathbf{V}_1 \text{diag}(\boldsymbol{\sigma}) \mathbf{V}_2^\top$ with singular values $\boldsymbol{\sigma}=[\sigma_1,\ldots,\sigma_{\text{min}(d_1,d_2)}]$ in descending order. $\mathbf{U}_1$ and $\mathbf{U}_2$ are computed from the top $k$ singular vectors of $\mathbf{T}$ as
$\mathbf{U}_1 = \boldsymbol{\Sigma}_{1}^{-1/2} \mathbf{V}_1^{(1:k)}$ and $\mathbf{U}_2 = \boldsymbol{\Sigma}_{2}^{-1/2} \mathbf{V}_2^{(1:k)}$ 
where $\mathbf{V}^{(1:k)}$ denotes the $k$ first columns of matrix $\mathbf{V}$. The sum correlation in the projection space is equivalent to 
\begin{equation}
  \sum_{i=1}^k \text{corr}\big(\big(\mathbf{u}_1^{(i)}\big)^\top X_1, \big(\mathbf{u}_2^{(i)})^\top \mathbf{X}_2 \big) = \sum_{i=1}^{k} \sigma_i^2\enspace,
  \label{eqn_cca_corr}
\end{equation}
i.e., the sum of the top $k$ squared singular values. 
\paragraph{Probabilistic CCA}
Latent variable model aims to learn a latent representation $\mathbf{z}\in\mathbb{R}^{N\times d_z}$ from a set of multivariate observations $\mathbf{X}\in\mathbb{R}^{N\times d_x}$ where $N$ is the number of observations while $d_z$ and $d_x$ are the dimensionality of the latent and observed data respectively. We denote the dimensions of the observed data as $\mathbf{X}_n$, $n\in[1,N]$ each consisting of $d$ features $\mathbf{X}_n \in \mathbb{R}^{d}$. The generative model specifies the relationship between the latent space and the observed as
\begin{align}
  \text{X}_{n} = f(\mathbf{z}_n) + \epsilon_{n},
\end{align}
where the form of the noise $\mathbf{\epsilon}$ leads to the likelihood function of the data.

The probabilistic CCA (PCCA) was proposed by \cite{Bach05,browne79} and it displays the shared latent representation explicitly. With a common vector of latent variables, $\mathbf{z} \in \mathbb{R}^{N\times d_z}$, $\mathbf{X}_1 \in \mathbb{R}^{N\times d_1}$ and $\mathbf{X}_2 \in \mathbb{R}^{N\times d_2}$ are modeled as
\begin{equation}
\begin{split}
  \mathbf{X}_1 &= \mathbf{W}_1 \mathbf{z} + \mathbf{\epsilon}_1, \notag \\
  \mathbf{X}_2 &= \mathbf{W}_2 \mathbf{z} + \mathbf{\epsilon}_2,\label{eq:ProbCCA}
\end{split}
\end{equation}
where $\mathbf{W}_1 \in \mathbb{R}^{d_1\times d_z}$, $\mathbf{W}_2 \in \mathbb{R}^{d_2\times d_z}$, the errors are distributed as $\epsilon_1 \sim \mathcal{N}_{d_1} (\mathbf{0}, {\mathbf{\Psi}}_1)$ and $\epsilon_2 \sim \mathcal{N}_{d_2} (\mathbf{0},{\mathbf{\Psi}}_2)$ where ${\mathbf{\Psi}_1}$ and
${\mathbf{\Psi}_2}$ are non-negative definite matrices, not
necessarily diagonal, allowing dependencies among the residual errors within an observation. 

PCCA defines the joint distribution over the random variables ($\mathbf{X}_1, \mathbf{X}_2$):
\begin{equation}
\begin{aligned}
 p(\mathbf{X}_1, \mathbf{X}_2,\mathbf{z}) &= p(\mathbf{z})p(\mathbf{X}_1|\mathbf{z})p(\mathbf{X}_2|\mathbf{z}), \\
 p(\mathbf{X}_1, \mathbf{X}_2) &=\int p(\mathbf{X}_1, \mathbf{X}_2,\mathbf{z}) d\mathbf{z}.
\end{aligned}
\end{equation}
The model assumes that $\mathbf{X}_1$ and $\mathbf{X}_2$ are independent conditioning on the latent variables $\mathbf{z}$. $p(\mathbf{X}_1|\mathbf{z})$ and $p(\mathbf{X}_2|\mathbf{z})$ are linear, however, linear observation models have limited representation power for complex data.

\section{Methods}
We now describe the proposed generalized DICCA for multi-view data. We assume that the $m$th view  $\mathbf{X}^{m} \in \mathbb{R}^{d_m\times 1}$ is independent with $N$ co-occurring observations. $\mathbf{Z}^m \in \mathbb{R}^{K\times 1}$ denote the $K$-dimensional latent representation specific to the $m$th view for  $m\in \{1,...,M\}$, where $M$ is the total number of views. $\mathbf{Z} \in \mathbb{R}^{K\times 1}$ denote  the $K$-dimensional latent representation common to all views. That is $\mathbf{Z}$ is the \textit{shared} latent variable capturing the  shared variation across $M$ views, while the \textit{view-specific} latent variables $\mathbf{Z}^m$ accounts for the view-specific variation. We then write the generative process of the latent variables as:
\begin{equation}
\begin{aligned}
\mathbf{Z}&\sim \mathcal {N} (\mathbf{0}_K,\mathbf{I}_K), \\ 
\mathbf{Z}^m &\sim \mathcal {N} (\mathbf{0}_K,\mathbf{I}_K).
\end{aligned}
\end{equation}

\textbf{View Generator} The variational autoencoder (VAE) \citep{Kingma14} propose the idea of amortized inference to perform variational inference in probabilistic models that are parameterized by deep neural networks. The limitation for deep generative models and VAE is that the learned representations are not easily interpretable due to complex interaction from latent dimensions to the observations. We consider view-specific generators and a linear latent-to-generator mapping with weights from a single latent dimension to a specific view. The view-sparse prior is applied over these grouped weights. We write the generative process of the data as:
\begin{align}
\label{eq:vae_factor_model}
	\mathbf{X}^{m} &\sim \mathcal{N} (f^{(m)}_{\theta_m} (\mathbf{\Lambda}^m\mathbf{Z}+ \mathbf{W}^m\mathbf{Z}^m), \mathbf{\Psi}^m),
\end{align}
where $\mathbf{\Lambda}^m,  \mathbf{W}^m \in \mathbb{R}^{d_m\times K}$.
The generator is encoded with the function $f^{(m)}_{\theta_m}(\cdot)$ specified as a deep neural network with parameters $\theta_m$, $\mathbf{\Psi}^m$ is a diagonal matrix containing the marginal variances of each component of $\mathbf{X}^m$. The latent representation $\mathbf{Z}$ is shared over all the view-specific generators, $\mathbf{Z}^m$ is view-specific. One of the main goals of this framework is to capture interpretable relationships between view-specific activations through the latent representation. 

\begin{figure}[ht]
\vskip 0.05in
\centering
\includegraphics[width=3.3in]{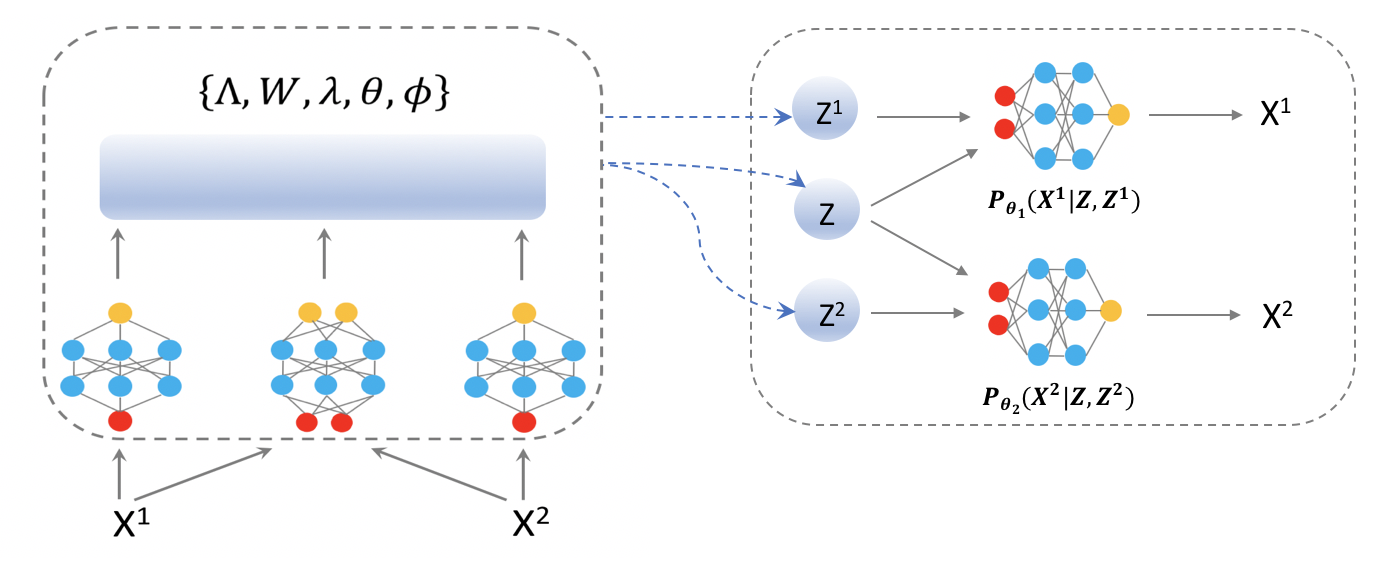}
\caption{Graphical illustration of the interpretable deep CCA model.}
\label{fig1}
\end{figure}

\textbf{Interpretable Sparsity Prior} $\mathbf{\Lambda}^{m}, \mathbf{W}^{m}$ are the \textit{latent-to-group matrices}. When the $j$th column of the latent-to-group matrix for view $m$, i.e., $\mathbf{\Lambda}^{(m)}_{:,j}, \mathbf{W}^{(m)}_{:,j}$ is all zeros, then the $j$th latent dimension, $\mathbf{z}_j$, will have no influence on view $m$.  To induce this column-wise sparsity, we place a hierarchical prior on the columns $\mathbf{\Lambda}^{(m)}_{:,j}, \mathbf{W}^{(m)}_{:,j}$ as follows \citep{Kyung10}:
\begin{equation}
\begin{aligned}
\gamma_{mj}^2 &\sim \text{Gamma}\left(\frac{d_m + 1}{2}, \frac{\lambda^2}{2}\right) \\
\mathbf{\Lambda}^{(m)}_{\boldsymbol{\cdot}, j}, \mathbf{W}^{(m)}_{\boldsymbol{\cdot}, j}  &\sim \mathcal{N}(\mathbf{0}, \gamma_{mj}^2 \mathbf{I}), 
\end{aligned}
\end{equation}
where Gamma($\cdot,\cdot$) is defined by shape and rate, and $d_m$ is the number of columns in each $\mathbf{\Lambda}^{(m)}, \mathbf{W}^{(m)}$. The rate parameter $\lambda$ defines the amount of sparsity, with larger $\lambda$ implying more column-wise sparsity in $\mathbf{\Lambda}^{(m)}$ and  $\mathbf{W}^{(m)}$. Marginalizing over $\gamma_{mj}^2$ induces view sparsity over the columns of $\mathbf{\Lambda}^{(m)}, \mathbf{W}^{(m)}$; the maximum a posterior estimator of the resulting posterior is equivalent to a group lasso penalized objective \citep{Kyung10}. Different from linear factor models, the deep structure of this model encourages the behavior to learn a set of $\mathbf{\Lambda}^{(m)}, \mathbf{W}^{(m)}$ matrices with very small weights only to have the values revived to ``appropriate'' magnitudes in the following layers of $f^{(m)}_{\theta_m}$. In order to mitigate such behavior a standard normal prior on the parameters of each generative network was placed, $\theta_m \sim \mathcal{N}(\mathbf{0}, \mathbf{I})$. Figure~\ref{fig1} provides a graphical illustration of our interpretable deep CCA model under $m$ = 2.

\subsection{Variational Inference} \label{sec:var_inf}
One unique feature of VAE \citep{Kingma14} is that it allows the conditional  $p(\mathbf{x} \mid \mathbf{z})$ being a potentially highly nonlinear
mapping from $\mathbf{z}$ to $\mathbf{x}$.

The likelihood is then parameterized with a generative network (called decoder). VAE uses $q(\mathbf{z}| \mathbf{x})$ with an inference network (called encoder) to approximate the posterior distribution of $\mathbf{z}$. For example, $q(\mathbf{z}|\mathbf{x})$ can be a Gaussian $\mathcal{N}(\mu, \sigma^2I)$, where both $\mu$ and $\sigma^2$ are parameterized by a neural network: $[\mu, \log \sigma^2] = f_\phi(\mathbf{x})$, where $f_\phi$ is a neural network with parameters $\phi$.
The parameters for both generative and inference networks are learned through variational inference, Jensen's inequality yields the evidence lower bound (ELBO) on the marginal likelihood of the data:

\begin{equation}
\begin{split}
\label{eqn:varlowbnd}
\log p_\theta(\mathbf{x})\ge
\underbrace{\mathbb{E}_{q(\mathbf{z}; \phi)}[\log p_\theta(\mathbf{x}\mid \mathbf{z})] - D_{KL}(q(\mathbf{z}; \phi)\mid\mid p(\mathbf{z})),}_{
\mathcal{L}(x; \theta, \phi)}
\end{split}
\end{equation}    
  
where $\mathrm{KL}(Q \| P)$ is Kullback-Leibler divergence between two distributions $Q$ and $P$. $q(\mathbf{z};\phi)$ is a tractable ``variational'' distribution meant to
approximate the intractable posterior distribution $p(\mathbf{z}\mid \mathbf{x})$; it is controlled by some parameters $\phi$.
We want to choose $\phi$ that makes the bound in Eq.~\eqref{eqn:varlowbnd} as tight
as possible.

One can train a feedforward
\emph{inference network} to find good variational parameters $\phi(\mathbf{x})$ for a given $\mathbf{x}$, where $\phi(\mathbf{x})$ is the output of a neural network with parameters $\phi$ that are trained
to maximize $\mathcal{L}(\mathbf{x}; \theta, \phi(\mathbf{x}))$~\citep{Kingma14}.

The \textit{KL divergence} between the approximate posterior and the prior distribution of the latent variables regularizes the prior knowledge about the latent variable for the learning algorithm.
The latent variables
$\mathbf{z}^m$ induced by the probabilistic graphical model of latent linear layer, we can write the approximate posterior of the set of latent variables as
$ q_{\eta}(\mathbf{z} | \mathbf{x}) = 
\prod_{m=1}^{M} q_{\eta}(\mathbf{z}_m | \mathbf{x})$
therefore, the KL divergence term can be decomposed to
\begin{align} \label{eqn:kl}
D_{KL} [ q_{\eta}(\mathbf{z} | \mathbf{x}) \Vert p(\mathbf{z}) ] = &
D_{KL} [ q_{\eta}(\mathbf{z} | \mathbf{x}) \Vert p(\mathbf{z}) ] +
\nonumber\\ &
\sum_{m=1}^{M} D_{KL} [ q_{\eta}(\mathbf{z}^m | \mathbf{x}^m) \Vert p(\mathbf{z}^m ) ]
\end{align}
More details can be found in appendix.

\subsection{Learning}
Traditionally, variational inference is learned by applying stochastic gradient methods directly to the evidence lower bound (ELBO) in equation \ref{eqn:varlowbnd}. We extend the basic amortized inference procedure to incorporate our sparsity inducing prior over the columns of the latent-to-group matrices. The naive approach of optimizing variational distributions for the $\gamma_{mj}^2$ and $\mathbf{W}_{\cdot,j}^{(m)}$ will not result in true sparsity of the columns $\mathbf{W}_{\cdot,j}^{(m)}$. Instead, we consider a collapsed variational objective function. Since our sparsity inducing prior over $\mathbf{W}_{\cdot,j}^{(m)}$ is marginally equivalent to the convex group lasso penalty we can use proximal gradient descent on the collapsed objective and obtain true group sparsity \citep{parikh14}.
Following the standard VAE approach of \citep{Kingma14}, we use simple point estimates for the variational distributions on the neural network parameters $\mathcal{W} = \left(\mathbf{W}^{(1)}, \cdots, \mathbf{W}^{(m)}\right)$, $\Lambda = \left(\mathbf{\Lambda}^{(1)}, \cdots, \mathbf{\Lambda}^{(m)}\right)$, $\theta = \left(\theta_1,\ldots,\theta_m\right)$, and  
We take $q_\phi(\mathbf{z}|\mathbf{x}) = \mathcal{N}(\mu(\mathbf{x}), \sigma^2(\mathbf{x})))$ where the mean and variances are parameterized by an inference network with parameters $\phi$.

\textbf{The collapsed objective}
Under $m\in \{1,...,M\}$, the data likelihood is defined by
\begin{equation}
\begin{aligned} 
&p_{\mathbf{\theta}}(\mathbf{x}^1,...,\mathbf{x}^m,\mathbf{z},\mathbf{z}^1,...,\mathbf{z}^m)\\ &= p(\mathbf{z}) \prod_{m=1}^{M}p(\mathbf{z}^m)p_{\mathbf{\theta}}(\mathbf{x}^m|\mathbf{z},\mathbf{z}^m;\mathbf{\theta}_m)\\
\end{aligned}
\end{equation}

We construct a collapsed variational objective by marginalizing the $\gamma_{mj}^2$ to compute $\log p_{\theta}(\mathbf{x})$ as:  
\begin{equation}
\begin{aligned}\label{eqn:elbo}
&\log p_{\theta}(\mathbf{x})  =  \log \int p(\mathbf{x} | \mathbf{z}, \mathbf{z}^1,..., \mathbf{z}^m, \mathcal{W}, \Lambda, \theta)  p(\mathbf{z}) \\& \times \prod_{m=1}^{M} p(\mathbf{z}^m) p(\mathcal{W} | \gamma^2)\times p(\Lambda | \gamma^2) p(\gamma^2) p(\theta) \,d\gamma^2 \,dz ... \,dz^m  \\
& \geq \sum_{m=1}^{M} E_{q_\phi(\mathbf{z} | \mathbf{x}^m),q_\phi(\mathbf{z}^m | \mathbf{x}^m)}  \left[ \log p_{\theta}(x^m | \mathbf{z},\mathbf{z}^m, \mathcal{W},\Lambda, \theta_m) \right] \\
&-  D_{KL} (q_\phi(\mathbf{z} | \mathbf{x}^1,...,\mathbf{x}^m) || p(\mathbf{z})) \\ &-\sum_{m=1}^{M}  D_{KL} (q_\phi(\mathbf{z}^m | \mathbf{x}^m) || p(\mathbf{z}^m)) \\
&+ \sum_{m=1}^{M}\log p(\theta_m) - \lambda \sum_{m,j} || \mathbf{\Lambda}^{(m)}_{\boldsymbol{\cdot}, j} ||_2 - \lambda \sum_{m,j} || \mathbf{W}^{(m)}_{\boldsymbol{\cdot}, j} ||_2 \\
& = \mathcal{L}(\phi, \theta, \mathcal{W},\Lambda).
\end{aligned}
\end{equation}


The columns of the latent-to-group matrices $\mathbf{\Lambda}_{\cdot,j}^{(m)}$ appear in a 2-norm penalty in the new collapsed ELBO. This is exactly a group lasso penalty on the columns of $\mathbf{\Lambda}_{\cdot,j}^{(m)}$ and encourages the entire vector to be set to zero.

Now our goal becomes to maximize this collapsed ELBO over $\phi, \theta, \mathcal{W}, \Lambda$.
%
Since this objective contains a standard group lasso penalty, we can leverage efficient proximal gradient descent updates on the latent-to-group matrices $\mathcal{W}$. Proximal algorithms achieve better rates of convergence than sub-gradient methods and have shown great success in solving convex objectives with group lasso penalties.
We use Adam for the remaining neural net parameters, $\theta_m$ and $\phi$.  
The details for optimization are included in the Appendix.
\section{Related work}
Deep variational CCA (DVCCA) \citep{wang16}  is a  variational CCA for two-view data representation learning. DVCCA shows that by modeling the view-specific variables that are specific to each view, DVCCA can disentangle shared/private variables and provide higher-qualify features and reconstructions. 

Later on, a two-view deep probabilistic CCA (DPCCA) was proposed \citep{gundersen20} based on PCCA \citep{Bach05} and convolutional neural networks for paired medical images and gene expression levels. The estimation requires PCCA and neural networks work simultaneously which is the big limitation for reconstruction and that is why they perform worse than multimodal autoencoder (MAE) in their reconstruction experiment. Another limitation is that DPCCA is designed specifically for paired image data.

oi-VAE \citep{Ainsworth18} is the first generative nonlinear group latent factor model, it combines deep generative models with a hierarchical sparsity-inducing prior that leads to the ability to extract meaningful interpretations of latent-to-observed interactions when the observations are structured into groups. DLGFA \citep{qiu20} is an advanced temporal extension of oi-VAE which can learn the dynamic dependency among groups through the shared latent variable and disentangle the interpretable dynamics among groups. Both oi-VAE and DLGFA use multiple decoders over the same latent variables, with the goal of having interpretable factors for the multi-view data. Compared to DVCCA, oi-VAE and DLGFA are interpretable, however, they do not model the common and view-specific variations like DVCCA which leaves a challenge for modeling complex multi-view data where the view-specific variations is hard to model. VPCCA \citep{karami21} can disentangle shared and view-specific variations and can generalize to multiple views, however, VPCCA lacks model interpretation. Additionally, the existing model interpretations are based on latent factor level not feature level which also poses a big application limitation for high-dimensional problems, like multi-omics data, where researchers are more interested in particular features. 

Based on the existing methods, our model is built on the variational generative model framework for efficient approximation purpose and we jointly model share and view-specific variations for complex multi-view data. Most importantly, we place sparsity-inducing prior on the latent weights to enable us achieve feature level interpretability.

\section{Experiments}
We empirically evaluate the representation learning, classification, and model interpretability of the proposed method. The performances of the proposed method are evaluated over the following three real-world datasets. The model architecture and implementation details are included in the Appendix.
\subsection{Noisy MNIST dataset}
Two-view noisy MNIST datasets from \citep{wang-multi-view15,wang16} are widely used in recent multi-view models.
The first view of the dataset is generated by rotating each image at angles randomly sampled from uniform distribution $\mathcal{U}(-\pi/4, \pi/4)$, while the second view is from randomly sampled images with same identity to the first view but not necessarily the same image and then is corrupted by random uniform noise. 
Both views thus share the same identity of the digit but not the style of the handwriting in the same class. The original training set is split into training/tuning sets of size 50$K$/10$K$. The data generation process ensures that the digit identity is the only common variable underlying both views. The performance is measured on the 10$K$ images in the test set. We follow the same neural network architecture that used in \citep{wang-multi-view15} and \citep{wang16} to make a fair comparison.  All the inference networks and decoding networks are composed of 3 fully connected nonlinear hidden layers of size 1024 units, where \text{ReLU} gate is used as nonlinearity for all the hidden units of the deep networks. We tune the latent dimension $K$ over [10,20,30,40,50], and fix $K_1 = K_2 = 30$, $K_1$ and $K_2$ represent the latent dimensions for the two dataset,  $\lambda=1$ based on the mean squared error on the test dataset.

\begin{table}[ht]

\caption{Reconstruction comparison on noisy two-view MNIST} 
\begin{center}
\begin{tabular}{l|c|c}
\hline
	Method 									 &\small{View 1 MSE (STD) }   &\small{View 2 MSE (STD) }  \\
	 \hline 
	\textbf{oi-VAE }		& 	0.059 (0.009) &                 0.172 (0.009)                                 \\ \hline
	\textbf{DPCCA }				&  0.052 (0.012) &                 0.134 (0.003)                                      \\ \hline
	
	\textbf{VCCA}							& 0.023 (0.011) &                 0.088 (0.0042)                                                   \\ \hline
	\textbf{VCCA-p }				& 0.024 (0.011) &                 0.084 (0.005)                                                   \\ \hline 
	\textbf{DICCA (Ours)}				& \textbf{0.016 (0.005)} &             \textbf{0.080 (0.005)}                                   \\ 
	\hline

\end{tabular}
\end{center}
\label{t:reconstruction}
\end{table}

\begin{figure*}[ht]
\vskip -0.05in
\centering
\includegraphics[width=6.2in]{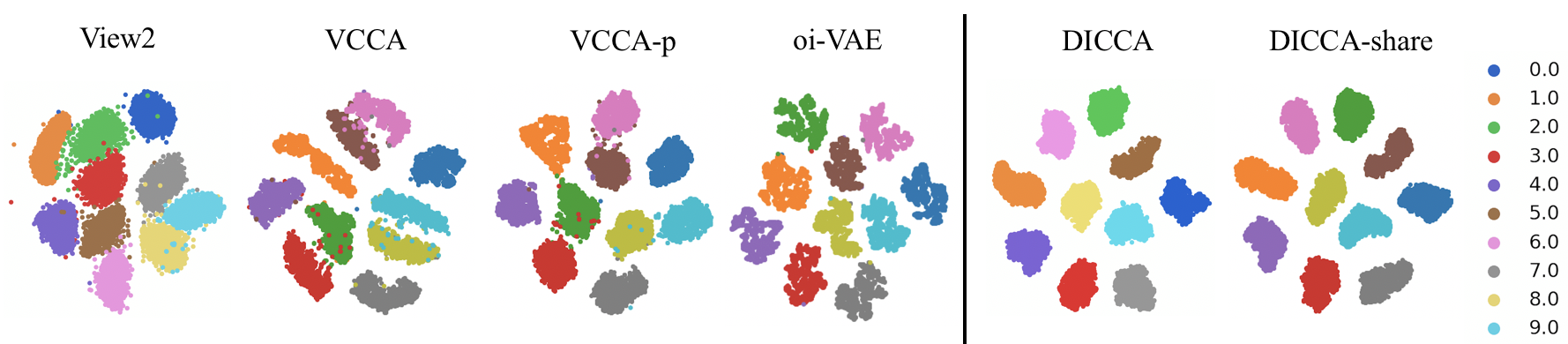}
\caption{t-SNE visualization of the extracted latent variable ${\bf z}$ from  images of view 2 on noisy MNIST test set by VCCA, VCCA-p, oi-VAE, DICCA, and DICCA-share.}
\label{fig:tsne}
\end{figure*}

\begin{figure}[th]
\vskip -0.05in
\centering
\includegraphics[width=3.4in]{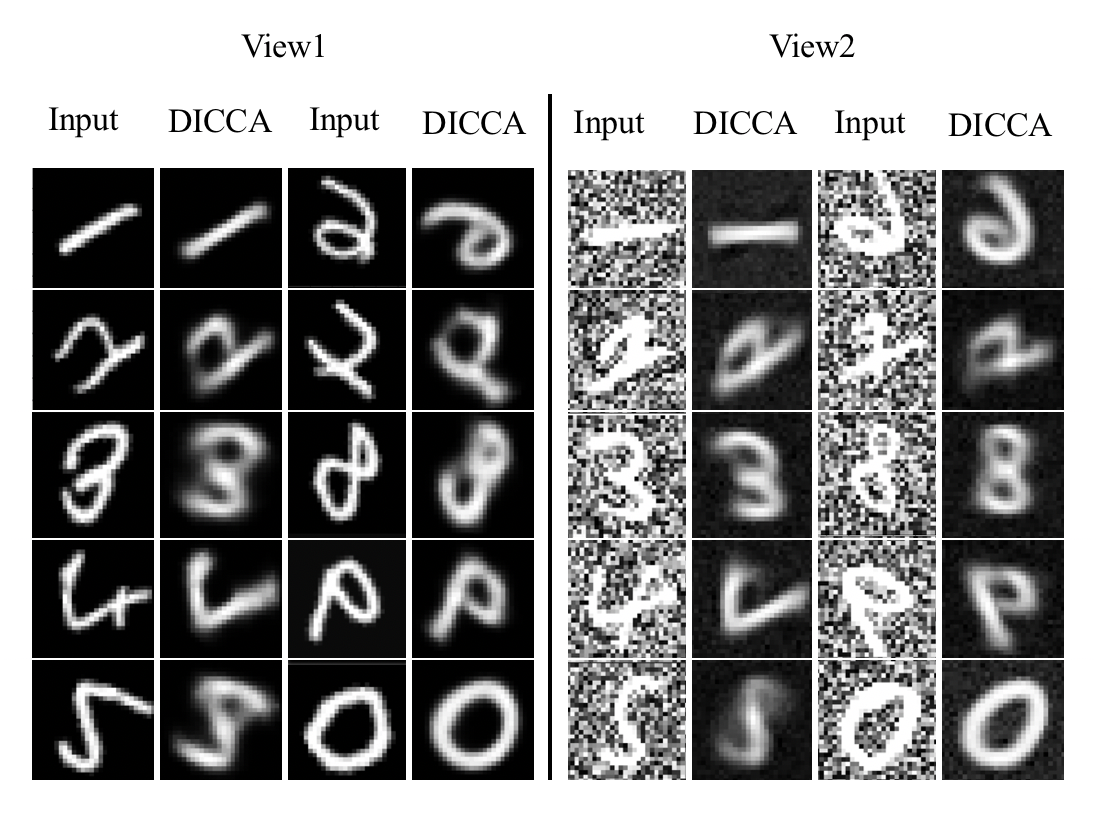}
\caption{Reconstruction of images from the noisy MNIST test set by DICCA. }
\label{fig:minst}
\end{figure}

{\bf Disentanglement learning} To evaluate the learned representation, our model should be able to reconstruct both views using the shared and view-specific latent variables. As baseline, we fit oi-VAE \citep{Ainsworth18}, DPCCA \citep{gundersen20}, VCCA \citep{wang16} to both data views. VCCA-p represents VCCA-private.  We find that DICCA can reconstruct both views well relative to these baselines (Table \ref{t:reconstruction}). oi-VAE performs the worst since it does not model the view-specific variations, DPCCA also does not perform  well because it requires optimizing PCCA in an inner loop. VCCA and VCCA-p perform much better than oi-VAE and DPCCA, but they are worse than DICCA which indicates that modeling view-specific variations is more powerful to extract the hidden truth than  modeling only the common variations. DICCA performs better in reconstruction than VCCA and VCCA-p, this confirms that the view-generator structure of DICCA is more suitable for learning multi-view data rather than using a single encoder of VCCA and VCCA-P. Figure \ref{fig:minst} shows sample reconstruction of noisy MNIST dataset by DICCA for view 1 (left) and view 2 (right). We can see that DICCA can capture the styles of each image very well and it can separate the background noise from the view 2 images. Additionally, in Figure \ref{fig:tsne} we provide 2D $t$-SNE  embedding of the view-specific latent representations from view 2 images learned by VCCA, VCCA-p, oi-VAE, DICCA, and DICCA-share (the shared latent projections). All the methods show improved separation performance compared to the original input data that we use for $t$-SNE. VCCA and VCCA-p perform similarly, but, there are some digits not very well separated, e.g.,  digit 3 and digit 2, digit 6 and digit 5. We also observe that oi-VAE  has the similar problem. This indicates that oi-VAE cannot capture the view-specific variations. DICCA and DICCA-share perform surprisingly well which quantitatively verify that the learned features of the images of different classes are well separated by view-specific latent variables and the shared latent projections.

{\bf Discriminative task} For the discriminative task, we apply SVM classification algorithm on the shared representation. We tune the parameters of SVM algorithm using the validation set, and the classification error is measured on the test set. The performance measures are AC (clustering accuracy rate), normalized mutual information (NMI), and classification error rate of a linear SVM \citep{cai05}. The baseline methods are Linear CCA: linear single layer CCA, DCCA: deep CCA \citep{andrew13}, Randomized KCCA: randomized kernel CCA approximation with Gaussian RBF kernels and random Fourier features \citep{lopez-paz14}, DCCAE: deep CCA auto-encoder \citep{wang-multi-view15}, VCCA: multi-view variational auto-encoder \citep{wang16}, and its shared-private multi-view variational auto-encoder. The results of the baselines are from \citep{wang-multi-view15, wang16}.

From table \ref{t:mnist_acc_nmi}, DICCA significantly improves the representation learning and the downstream classification accuracy.

\begin{table}[t]
\centering
\caption{Performance of several representation learning methods on the noisy MNIST digits test set.  Performance measures are clustering accuracy (AC), normalized mutual information (NMI) of clustering, and classification error rates of a linear SVM on the projections.}
\label{t:mnist_acc_nmi}
\begin{tabular}{l|c|c|c}
\hline
Method & AC (\%) & NMI (\%) & Error (\%) \\
\hline
\textbf{CCA}\  & 72.9 & 56.0 & 19.6 \\
\textbf{SVAE}\  & 64.0 & 69.0 & 11.9 \\
\textbf{KCCA}  & 94.7 & 87.3 & \hspace{.4em} 5.1 \\
\textbf{DCCA}\  & 97.0 & 92.0 & \hspace{.4em} 2.9 \\
\textbf{DCCAE}\  & 97.5 & 93.4 & \hspace{.4em} 2.2 \\
\textbf{VCCA}\  & 97.0 & 92.1 & \hspace{.4em} 3.0 \\
\textbf{VCCA-p}\  & 97.3  &  92.5 & \hspace{.4em} 2.4 \\
\textbf{DICCA (Ours)}\  & \textbf{98.0} & \textbf{94.0} & \hspace{.4em} \textbf{1.6} \\
\hline
\end{tabular}
\end{table}
\subsection{Drug biomarker discovery}
Chronic lymphocytic leukaemia (CLL), which combined $\textit{ex vivo}$ drug response measurements with somatic mutation status, transcriptome profiling and DNA methylation assays \citep{dietrich18}. There are four measurements on the same patients ($N=200$), in which mutations ($D=69$), mRNA($D=5000$), Methylation($D=4248$), and Drug response ($D=310$). We applied DICCA on CLL dataset to show model interpretability by exploring group dependency relationship and latent dimensions' interpretation and annotation. We use $K=10$ here after tuning. 

\begin{figure*}[ht]
\vskip -0.05in
\centering
\includegraphics[width=4.5in]{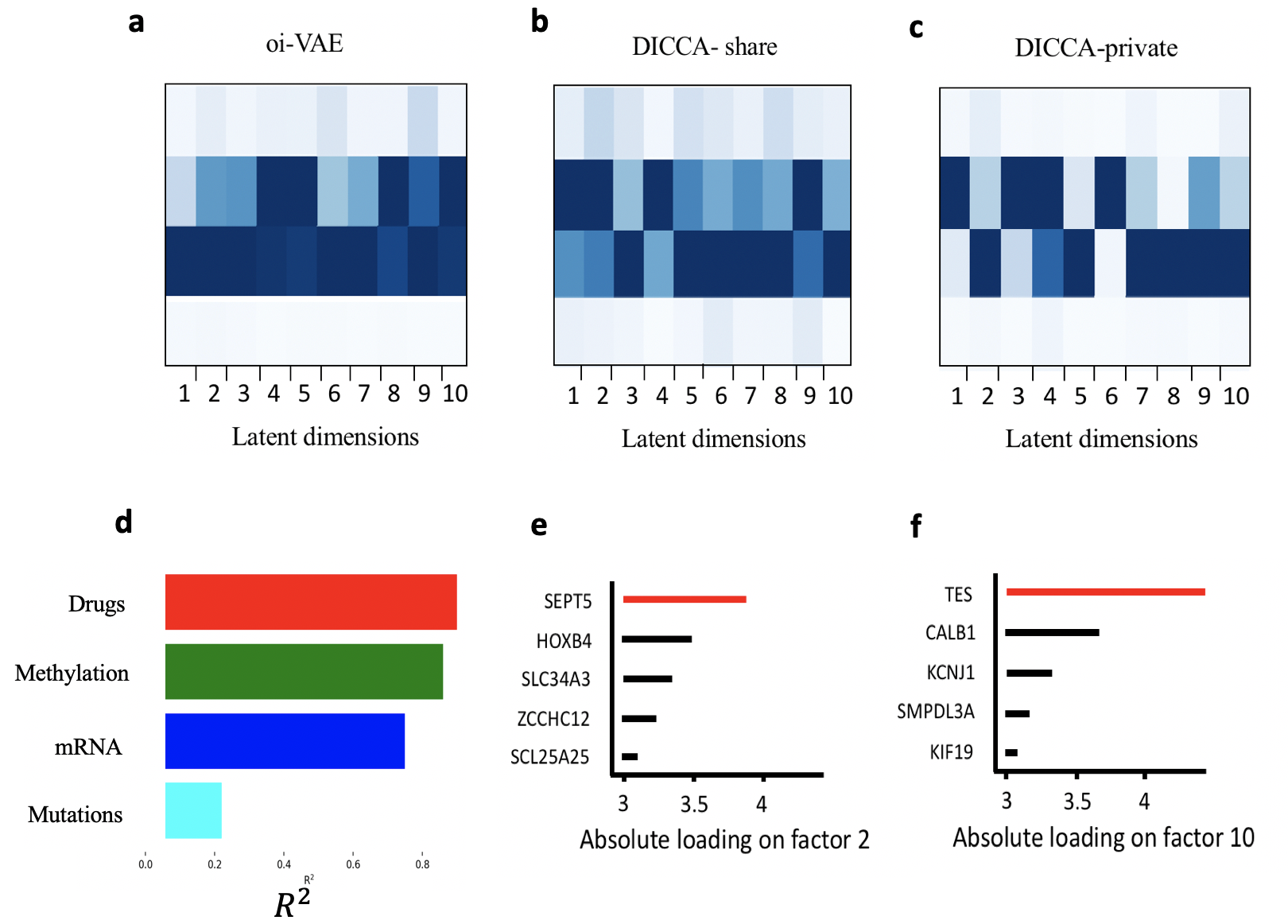}
\caption{Results on CLL data.  a-c: The learned $\mathbf{W}_{.,j}^{(m)}$ from oi-VAE, the learned $\mathbf{\Lambda}_{.,j}^{(m)}$ from DICCA, name as DICCA-share, the learned $\mathbf{W}_{.,j}^{(m)}$ from DICCA, name as DICCA-private. Specifically, the values of latent dimensions are color-coded from white (zero) to dark blue (maximum non-zero value) to indicate the strength of the latent-to-group mappings. d: cumulative proportion of total variance explained ($R^2$) by each view. e-f: Absolute loadings of top features of latent dimensions 2 and 10 in the mRNA data, top feature is marked as red color. }
\label{fig:cll}
\end{figure*}

{\bf Group dependency relationship} Each latent dimension of $\mathbf{z}$ influences only a sparse subset of the observational groups. We can view the observational groups associated with a specific latent dimension. To check the latent weight matrix can give us a bipartite graph in which we can quickly identify correlation and independence relationships among the groups themselves. This group dependency correlation among multi-view data is attractive as an exploratory tool independent of building a generative model. In our case, we will explore the shared and view-specific dependency respectively. We compare the group dependency extracted by oi-VAE \citep{Ainsworth18}, DICCA-share and DICCA-private in Figure \ref{fig:cll} a, b, and c. Both oi-VAE and DICCA show that methylation and mRNA data have the dominant variations across the 10 latent dimensions. As expected, we observe the view-specific variations are captured by DICCA-private, for example, under dimension 6, in oi-VAE the dominant variation is explained by mRNA which is also true in DICCA-share, however, in DICCA-private the variations are explained by methylation. Another example is under dimension 2, oi-VAE shows the dominant variations are from the mRNA, however, in DICCA-share we observe that the shared dominant variations are from methylation, but, the DICCA-private shows the view-specific variations are from mRNA.  

{\bf Latent dimension interpretation and biomarker discovery}\label{method:annotation} After the model has been trained, the first step is to disentangle the variations in each view. We compute the fraction of the variance explained ($R^2$) per view by
$$R_m^2 = 1 - \frac{(\sum_{n,d}X_{n,d}^m - \sum_{k}z_{nk}W_{kd}^m - \sum_{k}z_{nk}^mW_{kd}^m)^2}{(\sum_{n,d}X_{n,d}^m)^2}. $$
Subsequently, each dimension is characterized by two complementary analyses:
\begin{itemize}
    \item Ordination of the samples in factor space: Visualize a low dimensional representation of the main drivers of sample heterogeneity.
    \item Inspection of top features with largest weight: The loadings can give insights into the biological process underlying the heterogeneity captured by a latent dimension. We scale each weight vector by its absolute value.
\end{itemize}

In Figure \ref{fig:cll} d, we plot the variance explained by each view, DICCA explained 90\%, 86\%, 75\% variations in drug, methylation, mRNA, and only 22\% in mutations. This is much higher compared to MOFA \citep{Argelaguet18}. Based on the top weights in mRNA data, factor 2 was aligned with SEPT5 which is a member of the septin gene family of nucleotide binding proteins. Disruption of septin functions disturbs cytokinesis and results in large multinucleate or polyploid cells \citep{elzamly18}. Cancer-associated chromosomal changes often involve regions containing fragile sites. Factor 10 was aligned with TES which maps to a commom fragile site on chromosome 7q31.2 designated FRA7G. The TES gene lies within the minimal region of overlap of several LOH studies and appears to possess the properties of a tumour suppressor. TES is a negative regulator of cell growth and  may act as a tumour suppressor gene that is inactivated primarily by transcriptional silencing resulting from CpG island methylation \citep{tobias01}.

\subsection{Single-cell multi-omics study}
\begin{figure}[ht]
\vskip 0.05in
\centering
\includegraphics[width=3.3in]{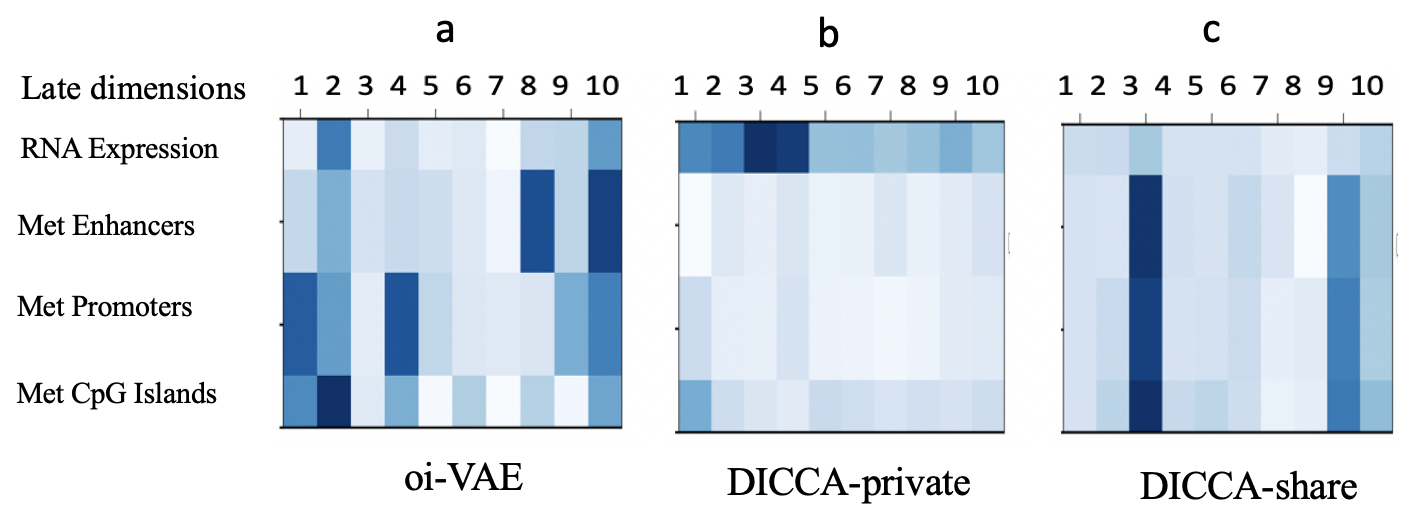}
\caption{Results on single-cell multi-omics data.  a-c: The learned $\mathbf{W}_{.,j}^{(m)}$ from oi-VAE, the learned $\mathbf{\Lambda}_{.,j}^{(m)}$ from DICCA, name as DICCA-share, the learned $\mathbf{W}_{.,j}^{(m)}$ from DICCA, name as DICCA-private. The values of latent dimensions are color-coded from white (zero) to dark blue (maximum non-zero value) to indicate the strength of the latent-to-group mappings. }
\label{fig:scmt}
\end{figure}

Multimodal methods are emerging in single cell biology \citep{clark18}. We applied DICCA and oi-VAE \citep{Ainsworth18} to disentangle the heterogeneity observed in a dataset of 87 ($N=87$) mouse embryonic stem cells (mESCs) \citep{angermueller16}, in which all the cells were profiled using single-cell methylation and transcriptome sequencing. This data contains transcriptome (RNA expression, $D=5000$) and the CpG methylation at three different genomic contexts: promoters ($D=5000$), CpG islands ($D=5000$) and enhancers ($D=5000$). In previous studies, MOFA \citep{Argelaguet18} identified three major factors driving cell-cell heterogeneity. DICCA-share in Figure \ref{fig:scmt} c is aligned with MOFA's findings that there is one latent dimension (factor 3 in DICCA-share) shared across all data modalities and the methylation show the dominant variations. However, the results from oi-VAE are different with MOFA and ours which no dimensions show the dominant variations from methylation. Additionally, both MOFA and oi-VAE can only model the shared variations among data, from DICCA-private in Figure \ref{fig:scmt} b, it show that the RNA expression is the dominant variation source under all dimensions, which means RNA expression data has large variations and is very noisy. The results from oi-VAE shows RNA expression is associated with methylation under factor 2 and factor 10, but it is difficult to interpret what factor 2 and factor 10 are related with. In real situations, this factor level interpretation has very limited understanding for the biological data. On the contrary, by further checking the latent weights of DICCA, the findings from the private $\mathbf{\Lambda}_{.,j}^{(m)}$ can give the researchers more biological insight on the feature level as described in CLL study for latent dimension interpretation and biomarker discovery.

\section{Discussion}
In this work, we develop a deep interpretable variational canonical correlation analysis for multi-view learning. 
It has been shown that following the view-generator and group sparsity formulation of the linear latent CCA model, we can obtain an interpretable learning algorithm for multi-view data. Empirical results have shown that this can efficiently disentangle the relationship among multiple views to obtain a more powerful representation which achieved state-of-the-arts performance on several downstream tasks. Besides the outperformed representation learning achieved by jointly modeling the share and view-specific variations, the proposed method can also have better interpretations for the latent dimensions. 

\bibliography{main}
\clearpage
\subsection*{A. Supplementary Methods}
\textbf{A.1. Proof of additive property of KL in Equation 9.} 

The approximate posterior of the set of latent variables can be factorized as 

\begin{equation}
q_{\eta}(\mathbf{z} \mid \mathbf{x})=q_{\eta}(\mathbf{z} \mid \mathbf{x}) \prod_{m=1}^{M} q_{\eta}\left(\mathbf{z}^m \mid \mathbf{z}, \mathbf{x}\right).
\end{equation}

We assume independent prior distribution prior on the latent variables which leads to

\begin{equation}
\begin{split}
&D_{\mathrm{KL}}\left[q_{\eta}(\mathbf{z} \mid \mathbf{x}) \| p(\mathbf{z})\right]=\int q_{\eta}(\mathbf{z} \mid \mathbf{x}) \prod_{m=1}^{M} q_{\eta}\left(\mathbf{z}^m \mid \mathbf{z}, \mathbf{x}\right) \\
&\times \log \frac{q_{\eta}(\mathbf{z} \mid \mathbf{x}) \prod_{m=1}^{M} q_{\eta}\left(\mathbf{z}^m \mid \mathbf{z}, \mathbf{x}\right)}{p(\mathbf{z}) \prod_{m=1}^{M} p\left(\mathbf{z}^m\right)} \\
&=\int q_{\eta}(\mathbf{z} \mid \mathbf{x}) \log \frac{q_{\eta}(\mathbf{z} \mid \mathbf{x})}{p(\mathbf{z})} \\
& +\sum_{m=1}^{M} \int q_{\eta}\left(\mathbf{z}^m \mid \mathbf{z}, \mathbf{x}\right) \log \frac{q_{\eta}\left(\mathbf{z}^m \mid \mathbf{z}, \mathbf{x}\right)}{p\left(\mathbf{z}^m\right)} \\
&=D_{\mathrm{KL}}\left[q_{\eta}(\mathbf{z} \mid \mathbf{x}) \| p(\mathbf{z})\right]+\sum_{m=1}^{M} D_{\mathrm{KL}}\left[q_{\eta}\left(\mathbf{z}^m \mid \mathbf{x}\right) \| p\left(\mathbf{z}^m\right)\right].
\end{split}
\end{equation}

\textbf{A.2. Proximal gradient descent} 

 A \textit{proximal algorithm} is an algorithm for solving a convex optimization problem which uses the proximal operators of the objective terms. Consider the problem
\begin{align}
\min_{x} f(x) + g(x),
\end{align}
where $f:\textbf{R}^n \rightarrow \textbf{R}$ and $g:\textbf{R}^n \rightarrow \textbf{R} \cup \{+\infty\}$ are closed proper convex and $f$ is differentiable.

The proximal gradient method is\begin{align}
\label{eq:proximal_step}
x^{k+1} = \text{prox}_{\lambda^k g} 
\end{align}
where $\lambda^k > 0$ is a step size, $\text{prox}_f (x)$ is the proximal operator for the function $f$. Expanding the definition of $ \text{prox}_{\lambda^k g}$, we can show that the proximal step corresponds to minimizing $g(x)$ plus a quadratic approximation to $g(x)$ centered on $x^k$. For $g(x) = \eta ||x||_2$, the proximal operator is given by
\begin{align}
\text{prox}_{\lambda^k g} (x) = \frac{x}{||x||_2} \left(||x||_2 -  \lambda^k \eta \right)_+ 
\end{align}

According to Parikh and Boyd, we know $(v)_+ \overset{\Delta}{=} \max (0, v)$. This operator can reduce $x$ by $\lambda \eta$, and $x$ can be shrinked to zero under $||x||_2 \leq \lambda^k \eta $. oi-VAE used proximal stochastic gradient updates for $\mathcal{W}$ and found that collapsed variational inference with proximal updates can not only provided faster convergence but also achieved model sparsity, we apply proximal stochastic gradient updates on $\mathcal{W}$ matrices and Adam on the remaining parameters. 

\subsection*{B. Model architecture and training procedure}

\textbf{Selection on $\lambda$ and $k$}
The parameter $\lambda$ control the model sparsity, larger $\lambda$ will imply more column-wise sparsity in $\mathbf{W}_{\boldsymbol{:},j}^{(m)}$, we propose to select $\lambda$ based on the learned $\mathbf{W}_{\boldsymbol{:},j}^{(m)}$ to check the sparsity and the MSE[test]. The latent dimension $k$ is chosen based on interpretation purpose.
\\

\textbf{B.1. Two-view noisy MNIST experiments} \\

We have view-specific encoder for each view, $\text{Encoder}_1$,  $\text{Encoder}_2$  and shared encoder $\text{Encoder}_{share}$ . After tuning, we use $d_1$ = $d_2$ = $d_{share}$ = 30.
\begin{itemize}
    \item $\text{Encoder}_1$:\\
    - $\mu(\mathbf{x_1}) = \mathbf{W}_1\text{relu} (\mathbf{x_1}) + b_1$.\\
    - $\sigma(\mathbf{x_1}) = \mathbf{W}_2 \text{softplus} (\mathbf{x_1}) + b_2$.\\
    
     \item $\text{Encoder}_2$:\\
     - $\mu(\mathbf{x_2}) = \mathbf{W}_3\text{relu} (\mathbf{x_2}) + b_3$.\\
    - $\sigma(\mathbf{x_2}) = \mathbf{W}_4 \text{softplus} (\mathbf{x_2}) + b_4$.\\
    
      \item $\text{Encoder}_{share}$:\\
    - $\mu(\mathbf{x_1 + x_2}) = \mathbf{W}_5 (\mathbf{x_1 + x_2}) + b_5$.\\
    - $\sigma(\mathbf{x1 + x_2}) = \text{exp}(\mathbf{W}_6 (\mathbf{x_1 + x_2}) + b_6)$.\\
    
    \item Decoder:\\
    - $\mu(\mathbf{z}) = \mathbf{W_7} \text{tanh} (\mathbf{z}) + b_7$.\\
    - $\sigma(\mathbf{z}) = \text{exp}(b_8)$.\\
\end{itemize}
The learning rate on $\mathcal{W}$ is 1e-4 for encoder and decoder, batch size is 128. Optimization was run for 1,000 epochs.

\textbf{B.2. CLL experiments} \\

We have view-specific encoder for each view, $\text{Encoder}_{drug}$,  $\text{Encoder}_{methylation}$, $\text{Encoder}_{mRNA}$, $\text{Encoder}_{mutation}$ and shared encoder $\text{Encoder}_{share}$ . After tuning, we use $d_{drug}$ = $d_{methylation}$ = $d_{mRNA}$ = $d_{mutation}$ = $d_{share}$ = 10.
\begin{itemize}
    \item $\text{Encoder}_{drug}$:\\
    - $\mu(\mathbf{x_1}) = \mathbf{W}_1\text{relu} (\mathbf{x_1}) + b_1$.\\
    - $\sigma(\mathbf{x_1}) = \mathbf{W}_2 \text{softplus} (\mathbf{x_1}) + b_2$.\\
    
     \item $\text{Encoder}_{methylation}$:\\
     - $\mu(\mathbf{x_2}) = \mathbf{W}_3\text{relu} (\mathbf{x_2}) + b_3$.\\
    - $\sigma(\mathbf{x_2}) = \mathbf{W}_4 \text{softplus} (\mathbf{x_2}) + b_4$.\\
    
    \item $\text{Encoder}_{mRNA}$:\\
    - $\mu(\mathbf{x_3}) = \mathbf{W}_5\text{relu} (\mathbf{x_3}) + b_5$.\\
    - $\sigma(\mathbf{x_3}) = \mathbf{W}_6 \text{softplus} (\mathbf{x_3}) + b_6$.\\
    
     \item $\text{Encoder}_{mutation}$:\\
     - $\mu(\mathbf{x_4}) = \mathbf{W}_7\text{relu} (\mathbf{x_4}) + b_7$.\\
    - $\sigma(\mathbf{x_4}) = \mathbf{W}_8 \text{softplus} (\mathbf{x_4}) + b_8$.\\

      \item $\text{Encoder}_{share}$:\\
    - $\mu(\mathbf{x_1 + x_2 + x_3 + x_4}) \\ = \mathbf{W}_9 (\mathbf{x_1 + x_2 + x_3 + x_4}) + b_9$.\\
    - $\sigma(\mathbf{x_1 + x_2 + x_3 + x_4}) \\ = \text{exp}(\mathbf{W}_{10} (\mathbf{x_1 + x_2 + x_3 + x_4}) + b_{10})$.\\
    
    \item Decoder:\\
    - $\mu(\mathbf{z}) = \mathbf{W_{11}} \text{tanh} (\mathbf{z}) + b_{11}$.\\
    - $\sigma(\mathbf{z}) = \text{exp}(b_{12})$.\\
\end{itemize}
The learning rate on $\mathcal{W}$ is 1e-4 for encoder and decoder, batch size is 12. Optimization was run for 2,000 epochs.

\textbf{B.3. Single-cell multi-omics dataset} \\

The model architecture is similar as CLL study. We have view-specific encoder for each view, $\text{Encoder}_{RNA Expression}$,  $\text{Encoder}_{Met Enhancers}$, $\text{Encoder}_{Met Promoters}$, $\text{Encoder}_{Met CpG Islands}$ and shared encoder $\text{Encoder}_{share}$ . After tuning, we use $d_{RNA}$ = $d_{Enhancers}$ = $d_{Promoters}$ = $d_{CpG}$ = $d_{share}$ = 10.
The learning rate on $\mathcal{W}$ is 1e-4 for encoder and decoder, batch size is 10. Optimization was run for 2,000 epochs.
\end{document}